\documentclass[conference,a4paper]{IEEEtran}
\IEEEoverridecommandlockouts
\usepackage{cite}
\usepackage[T5]{fontenc}
\usepackage{algorithm}
\usepackage{algpseudocode}
\usepackage{amsmath,amssymb,amsfonts}
\usepackage{hyperref}
\usepackage{graphicx}
\usepackage{textcomp}
\usepackage{xcolor}
\def\BibTeX{{\rm B\kern-.05em{\sc i\kern-.025em b}\kern-.08em
    T\kern-.1667em\lower.7ex\hbox{E}\kern-.125emX}}
\begin{document}
\bstctlcite{bstctl:nodash}

\title{UIT-HWDB: Using Transferring Method to Construct A Novel Benchmark for Evaluating Unconstrained Handwriting Image Recognition in Vietnamese}

\author{Nghia Hieu Nguyen$^{1,2}$, Duong T.D. Vo$^{1,2}$, Kiet Van Nguyen$^{1,2}$ \\
$^{1}$University of Information Technology, Ho Chi Minh City, Vietnam \\
$^{2}$Vietnam National University, Ho Chi Minh City, Vietnam \\
Email: 19520178@gm.uit.edu.vn, 19520483@gm.uit.edu.vn, kietnv@uit.edu.vn}

\maketitle

\begin{abstract}
Recognizing handwriting images is challenging due to the vast variation in writing style across many people and distinct linguistic aspects of writing languages. In Vietnamese, besides the modern Latin characters, there are accent and letter marks together with characters that draw confusion to state-of-the-art handwriting recognition methods. Moreover, as a low-resource language, there are not many datasets for researching handwriting recognition in Vietnamese, which makes handwriting recognition in this language have a barrier for researchers to approach. Recent works evaluated offline handwriting recognition methods in Vietnamese using images from an online handwriting dataset constructed by connecting pen stroke coordinates without further processing. This approach obviously can not measure the ability of recognition methods effectively, as it is trivial and may be lack of features that are essential in offline handwriting images. Therefore, in this paper, we propose the Transferring method to construct a handwriting image dataset that associates crucial natural attributes required for offline handwriting images. Using our method, we provide a first high-quality synthetic dataset which is complex and natural for efficiently evaluating handwriting recognition methods. In addition, we conduct experiments with various state-of-the-art methods to figure out the challenge to reach the solution for handwriting recognition in Vietnamese.
\end{abstract}


\section{Introduction}
\IEEEPARstart{H}{andwriting} Recognition has been a challenging task in image processing and computational linguistics. This task is traditionally divided into two main categories: online handwriting recognition and offline handwriting recognition. Because of the linguistic aspect and low-resource status of Vietnamese, recognizing Vietnamese handwriting images is still an arduous task. As collecting handwriting images is problematic, there are not many large-scale and high-standard dataset for researching handwriting recognition in Vietnamese.

Previous works \cite{Shen20161,krishnan2016generating} attempted to synthetically conduct offline handwriting datasets by using handwriting fonts to render images. However, datasets constructed by these approaches is not guaranteed to be natural as (1) for the approach proposed in \cite{krishnan2016generating}, the stable form of computer fonts opposes to the various handwriting styles of human, which obviously makes font-rendered handwriting images have different distribution to realistic handwriting images; and (2) for the approach proposed in \cite{Shen20161}, the colors of strokes and background are not considered carefully, and re-arranging segmented characters to a defined line does not clearly express the unconstrained properties of realistic handwriting images. To this end, we analysed and conducted experiments and then propose a simple but effective method, called Transferring method, to construct an offline handwriting dataset from an online handwriting dataset. Using this method, we introduce the first offline handwriting dataset in Vietnamese in order to motivate research community to explore and conduct more experiments to ultimately create an effective framework to tackle offline handwriting recognition task in Vietnamese.
\section{Related works}
Marti et al. \cite{Marti2002TheIA} constructed the IAM dataset by creating forms containing computer-printed texts for volunteers to write down these texts in their own writing style. After collecting these filled forms, line segmentation algorithms and word segmentation algorithms were used to extract lines from handwritten texts and words from extracted handwritten lines. On the other hand, the RIMES dataset \cite{6065550} was constructed based on the content of mails. In both cases, a large number of annotators were called in to annotate the datasets, and the annotation processes were re-corrected many times for the quality assurance.


Motivated by previous works and the shortage of Vietnamese offline handwriting dataset, we propose a novel method, called Transferring method, to construct an offline handwriting dataset in Vietnamese by synthesizing fundamental attributes from different handwriting datasets. Our conducted experiments ensure the resulting dataset totally keep mainly essential properties of an offline handwriting image dataset, which bridge the gap between handwriting images captured in the real world and images constructed synthetically. Consequently, we introduce a novel offline handwriting image dataset in Vietnamese constructed using the Transferring method, named UIT-HWDB.
\section{Challenge from writing characteristics of Vietnamese language}
Like other Western languages, Vietnamese uses the modern Latin script for its writing system. However, there is a clear difference that Vietnamese relies heavily on diacritics, which makes Vietnamese handwriting more complicated in a distinct way. In the Vietnamese alphabet, there are seven letters that always have their attached diacritics: \emph{ă} (breve), \emph{â}, \emph{ê}, \emph{ô} (circumflex), \emph{ơ}, \emph{ư} (horn) and \emph{đ} (bar); and five additional diacritics used to designate tone: grave (as in \emph{à}), acute (as in \emph{á}), hook above (as in \emph{ả}), tilde (as in \emph{ã}), and dot below (as in \emph{ạ}) \cite{truong2015vietnamese}. When being scrawled, some diacritics may be mistakenly read or seen as others, especially for computer vision systems. Specifically, a hook above glyph may look like a horn glyph (\emph{ủ} versus \emph{ư}), an acute glyph may be carelessly written to be like a grave glyph (\emph{á} versus \emph{à}), and a dot below under a letter \emph{i} may make the character \emph{ị} look like an exclamation mark (\emph{ị} versus \emph{!}). The examples above, and many more mistakes caused by carelessly written diacritics, frequently happen in real life and be an obstacle for optical character recognition for Vietnamese handwriting. Those flaws are also what we have encountered in our research, and we will analyze how the models stumble through them in Section 5. 
\section{UIT-HWDB Dataset}
\subsection{Transferring method}
We present in this section the Transferring method to synthetically construct a Vietnamese handwriting image dataset. The novel constructed dataset must have the fully essential properties of a realistic handwriting image dataset. To achieve this requirement, the Transferring method first constructs the handwritten characters into an image, then applies the color variation to make the image more natural. The way of considering these real-world factors is detailed as follows:

\begin{algorithm}
    \caption{Pseudocode for the Transferring method ($U(a, b)$ is the discrete uniform distribution in range $[a, b]$, $Beta(\alpha, \beta)$ is the beta distribution).}\label{alg:cap}
    \begin{algorithmic}
        \Require \\
            \begin{itemize}
                \item $coords = ({x_n, y_n})_{n \in \mathbb{N}}$: the sequence of coordinates of a pen stroke.
                \item $all\_strokes$: the sequence of coordinates sequences of all strokes in the image.
                \item $bg\_dis = ({\alpha_m, \beta_m})_{m \in \mathbb{N}}$: the sequence of $m$ pairs of parameters for approximated beta distributions of background color.
                \item $stroke\_dis = ({\alpha_m, \beta_m})_{m \in \mathbb{N}}$: the sequence of $m$ pairs of parameters for approximated beta distributions of stroke color.
            \end{itemize}
        
        \Ensure a handwriting image containing characters drawed by the input coordinates.
        \\
        \Procedure{render\_image}{$all\_strokes$}
            \State $image \gets 255$ for all pixels
            \For{$coords$ in $all\_strokes$}
                \For{$ith$ in $[1:length(coords)]$}
                    \State $image[coords_{ith-1}:coords_{ith}] \gets 0$
                \EndFor
            \EndFor
            \State \textbf{return} image
        \EndProcedure
        \\
        \Procedure{render\_color}{image}
            \State $stroke\_ps \gets []$ \Comment{Sequence of stroke color variables}
            \State $bg\_ps \gets []$ \Comment{Sequence of background color variables}
            \For{$ith$ in [1:m]}
                \State $\alpha, \beta \gets stroke\_dis[ith]$
                \State $stroke\_ps.append(X \sim Beta(\alpha, \beta))$
            \EndFor
            \For{$ith$ in [1:m]}
                \State $\alpha, \beta \gets bg\_dis[ith]$
                \State $bg\_ps.append(X \sim Beta(\alpha, \beta)$)
            \EndFor
            \State $ith \sim U(1, m)$ \Comment{Randomly use $ith$ random variable, regarding both stroke and background color}
            \\
            \For{each pixel $p$ in image}
                \If{$p$ is 0} \Comment{Is stroke pixel}
                    \State $p \gets stroke\_ps[ith] \times 255.0$
                \Else \Comment{Is background pixel}
                    \State $p \gets bg\_ps[ith] \times 255.0$
                \EndIf
            \EndFor
        \EndProcedure
        \\

        \State $image \gets RENDER\_IMAGE(all\_strokes)$
        \State $image \gets RENDER\_COLOR(image)$
    \end{algorithmic}
\end{algorithm}

\subsubsection{Human handwriting style}\hspace{\fill} \\
The handwriting style is one of the specific properties that distinguish handwritten text from computer-printed text. Moreover, the variation of handwriting style is the most important characteristic required for a handwriting dataset, both online and offline.

Previous works \cite{krishnan2016generating} automatically used handwriting fonts to construct offline handwriting images, which causes the gap between natural handwriting images and synthetic handwriting images. To bridge this gap, we inherit handwritten characters from an online handwriting dataset, the VNOnDB \cite{nguyen2018database}. This approach guarantees the vast variation in natural writing style of human.

\subsubsection{Color}\hspace{\fill} \\
Recent works \cite{le2019end,le2018recognizing,9335877} constructed handwriting images in Vietnamese by connecting coordinates of pen stroke in the VNOnDB dataset \cite{nguyen2018database}. They kept the ink color as black (0 for pixel value) and the background color as white (255 for pixel value). This approach is intuitively not natural. Ignoring the color factors also means that we unintentionally reduce the complexity of the image compared with an image captured in reality, which causes the gap between research and realistic applications. To address these downsides, we take into account the complexity of the color.

For more details, we analyze the colors from the IAM dataset \cite{Marti2002TheIA}. The IAM dataset was collected manually, containing the real-world variation of both stroke colors and background colors. Therefore, the distributions of color in this dataset are natural enough to bridge the gap between synthetic images and real-world captured images. In the IAM dataset, for each image subset we intend to sample the stroke color values and background color values to get the corresponding color sampling distributions. We find that the color distributions have a finite range from 0 to 255 and have variously flexible shapes and skewnesses. After visualizing the color sampling distributions and taking into account the aforementioned statement, we finally come up with approximating the beta distribution for stroke colors and background colors of each subset. We then calculate and have a list of $\alpha$ and $\beta$ shape parameters for the beta distribution of the stroke colors and background colors. We apply to the strokes and background of handwriting images with random colors that follow the stroke and background color beta distribution which were generated from randomly picked pairs of $\alpha$ and $\beta$ parameters from the list. In this way, we ensure the nature of colors for handwriting images created synthetically.


\subsubsection{Stroke-width}\hspace{\fill} \\
Although most synthetic handwriting datasets are constructed in various ways that resemble images captured in the real world, they still miss the stroke-width variation. But in our work, experiments show that this factor is not essential and can be slightly ignored to keep the Transferring method simple without reducing the complexity of the dataset. We will carefully analyze this statement in Section 5.

\subsubsection{Annotation}\hspace{\fill} \\
Ground truth texts for the dataset are inherited from the ground truth texts of the VNOnDB dataset \cite{nguyen2018database}. However, from VNOnDB, there are some incorrect labels for handwriting images. Therefore, we also re-correct their mislabeled ground truth texts during the image constructing process.

\subsection{UIT-HWDB dataset}
Using the Transferring method with the VNOnDB as the online handwriting dataset and the IAM as the source for extracting colors, we construct the first novel offline handwriting dataset, the UIT-HWDB dataset. In more detail, our dataset has two parts: UIT-HWDB-word (110,745 unconstrained handwritten-word images) and UIT-HWDB-line (7,273 unconstrained handwritten-line images).



\begin{table}[ht]
    \renewcommand{\arraystretch}{1.3}
    \caption{Statistical comparison between handwriting image datasets (Para. stands for Paragraph, * indicates dataset in English, ** indicates dataset in French, $^{+}$ indicates dataset in German, $^{-}$ indicates dataset in Latin, $^{++}$ indicates dataset in Vietnamese).}
    \label{tab:1a}
    \centering
    \begin{tabular}{lccc}
        \hline
        Dataset             & Words   & Lines  & Para. \\ \hline
        RIMES**               & 83,493  & 11,365 & 1,500 \\
        IAM*                & 115,320 & 13,353 & -     \\
        Washinton Database*         & 4,894   & 656    & -     \\
        Parzival Database$^{+}$   & 23,487  & 4,477  & -     \\
        Saint Gall Database$^{-}$ & 11,597  & 1,410  & -     \\
        UIT-HWDB (ours)$^{++}$     & 110,745 & 7,273  & - \\ \hline
    \end{tabular}
\end{table}

We additionally reconstruct the test sets for the UIT-HWDB dataset as the original test sets of the VNOnDB generally contain easy-to-read images, which can lead to over-confidence in the ability of handwriting recognition methods. Specifically, we manually select both easy-to-read and hard-to-read handwriting images for our test set to evaluate the ability of recognition methods on many levels. The dataset is available at \url{https://github.com/hieunghia-pat/UIT-HWDB-dataset} for research purposes.

\subsubsection{Human evaluation}\hspace{\fill} \\
Our Transferring method is a semi-automatic method, which is undoubtedly prone to potential errors while constructing images. To carefully evaluate the errors of our Transferring method, we randomly collect 1,200 images from the UIT-HWDB-word train set and 600 images from the UIT-HWDB-line train set, then invite four people to manually evaluate the images (300 word-level images and 150 line-level images for each person). For the evaluation metric, we use the Cohen's kappa. As our method keeps the handwriting style, background color and stroke color consistent, the main factor that can cause the rendered characters to be abnormal is the choice of stroke width (detailed in section 5). Therefore the guideline for the four people to evaluate the images is to observe whether the rendered characters are deformed because of the improperly applied stroke width (stroke-width error) or the handwriting process of the annotators (non-stroke-width error). Results show the annotators' agreement for non-stroke-width error with Cohen's kappa coefficient $\mathbb{\kappa} = 1$ and $0$ for stroke-width error on the three collected sets. Our Transferring method totally keeps the original shape of characters.
\section{Experiments}
\subsection{Baseline methods}
The early approach for offline handwriting recognition is the combination of Convolutional Neural Network (CNN) \cite{o2015introduction}, Recurrent Neural Network (RNN) \cite{sherstinsky2020fundamentals}, and Connectionist Temporal Classification (CTC) loss \cite{graves2006connectionist}, which we call CTC-based methods. \cite{michael2019evaluating} proposed another combination using attention module \cite{cho2014learning} in their RNN layers. To enhance the ability of attention mechanisms in handwriting recognition methods, \cite{kang2020pay} proposed transformer-based methods, which have inherited the recent success of the transformer architecture in Natural Language Processing (NLP). In this experiment, we use TransformerOCR \cite{feng2020scene} and NRTR \cite{8978180} as the transformer-based methods. For attention-based methods, we implement Attention-based Encoder Decoder (AED) \cite{le2018recognizing} and finally we use CRNN \cite{shi2016end} and a method proposed in \cite{8269951} (BiCRNN for short) as the CTC-based methods.

\subsection{Evaluation metric}
For the Optical Character Recognition task, Damerau–Levenshtein Distance is the widely-used metric for measuring the performance of recognition methods. In this work, we use both two instances of this distance: Character Error Rate (CER) and Word Error Rate (WER).

\subsection{Experiments on the stroke-width factor}

In this experiment, we implemented the baseline methods on the word-level and line-level images.


To carefully analyze the effect of stroke-width factor on the ability of current handwriting image recognition methods, we created two versions for the UIT-HWDB-line dataset and UIT-HWDB-word dataset, named line-v1 and line-v2 (word-v1 and word-v2, respectively). The first version of these two datasets contains handwriting images without any variation in stroke width, while those in the second version contain the variation in the width of handwritten strokes. line-v1 and line-v2 are used to investigate the ability of TransformerOCR (transformer-based method) and BiCRNN (CTC-based method), while word-v1 and word-v2 are used to analyze the ability of NRTR (transformer-based method) and CRNN (CTC-based method). Note that we padded blank pixels to images in all training schemes in order to let all images have the same size and keep their original scale.

To mimic the variation of stroke-width factor, we approximate the width of pen stroke following the assumption: stroke-width is thinner when the writers draw up than when they draw down. Therefore, based on the ratio between the difference of two horizontal coordinates $ |\Delta x|$ and the difference of two vertical coordinates $|\Delta y|$ of two adjacent points, we construct a function $w$ which approximates the stroke width of handwritten characters:
\begin{equation}
    w(\theta) = m \cdot d(\theta)
\end{equation}
\begin{equation}
    d(\theta) = \displaystyle\frac{1}{1+e^{\alpha\theta + \beta}}
\end{equation}
In this experiment, we set $ \alpha = -0.1$, $\beta = 1.13$, $\theta = \arctan{\displaystyle\frac{\Delta y}{\Delta x}}$ and $m$ is the maximum thickness value a stroke can have.

To construct images with constant stroke width, we set for $\theta$ a constant value which makes the function $d(\theta)$ close to 1.

In practice, we observed the proper range for getting $m$ is $[2, 5]$. For the UIT-HWDB dataset, we randomly selected a value for $m \sim U(2, 5)$ where $U(a,b)$ is the discrete uniform distribution in range $[a, b]$  for every image.

\begin{table}
    \renewcommand{\arraystretch}{1.3}
    \caption{Results of TransformerOCR and BiCRNN on two versions of UIT-HWDB-line dataset (the first two rows were trained on line-v1 and last two rows were trained on line-v2).}
    \label{tab:3}
    \centering
    \begin{tabular}{lcccc}
        \hline
                & \multicolumn{2}{c}{line-v1} & \multicolumn{2}{c}{line-v2} \\ \cline{2-5} 
                & CER          & WER          & CER          & WER          \\ \hline
        TransformerOCR & \textbf{11.59}        & \textbf{21.38}        & \textbf{11.42}        & \textbf{21.67}        \\
        BiCRNN & 12.06        & 31.84        & 12.28        & 31.49        \\ \hline
        TransformerOCR & 16.88        & 30.61        & 12.18        & \textbf{22.33}        \\
        BiCRNN & \textbf{11.50}        & \textbf{29.78}        & \textbf{10.77}        & 27.54        \\ \hline
    \end{tabular}
\end{table}

The experimental results are shown in Table \ref{tab:3}. According to Table \ref{tab:3}, when we trained TransformerOCR and BiCRNN on line-v1 and then evaluated these two models on both line-v1 and line-v2 test sets, we recognized they had approximately the same results. However, when we trained these models on line-v2 and then evaluated them on the line-v1 test set and line-v2 test set, they performed well on the line-v2 but yielded a few drawbacks on the line-v1. In conclusion, these results showed that with their strong CNN structures, TransformerOCR and BiCRNN, when being trained on version 1, can generalize well the pattern of handwritten text in images, hence they concurrently yielded the same results when predicting on the test set of version 2.

\begin{table}[!ht]
    \renewcommand{\arraystretch}{1.3}
    \caption{Results of NRTR and CRNN on the two versions of the UIT-HWDB-word dataset (the first two rows were trained on word-v1 and last two rows were trained on word-v2).}
    \label{tab:4}
    \centering
    \begin{tabular}{lcccc}
        \hline
                & \multicolumn{2}{c}{word-v1}    & \multicolumn{2}{c}{word-v2}    \\ \cline{2-5} 
                & CER           & WER            & CER           & WER            \\ \hline
        NRTR & \textbf{8.28} & 22.00 & \textbf{10.10}         & 25.48          \\
        CRNN & 9.64 & \textbf{20.27} & 10.74         & \textbf{23.26}          \\ \hline
        NRTR & 18.01         & 37.73          & \textbf{8.32} & 21.21 \\
        CRNN & \textbf{16.66}         & \textbf{31.93}          & 9.77 & \textbf{20.62} \\ \hline
    \end{tabular}
\end{table}

Nevertheless, coming from NRTR and CRNN, we obtained another behavior. As Table \ref{tab:4} indicates, these two models only performed well on the version of the dataset that they had been trained. These results show that CRNN and NRTR can catch the specific attributes of each version of the dataset; thus, they are quite sensitive when making a prediction on another version of the images.

Finishing these experiments, we conclude that with current deep learning methods, stroke-width does not participate in forming the complexity of offline handwriting recognition task. Therefore we can ignore this factor to keep our Transferring method simple without lacking the nature of a handwriting image dataset.

\subsection{Experiments on baseline methods}
\begin{table}[!ht]
    \renewcommand{\arraystretch}{1.3}
    \caption{Results of baseline methods on UIT-HWDB-word and UIT-HWDB-line test set.}
    \label{table:2}
    \centering
    \begin{tabular}{l|cc|cc}
        \cline{1-5}
                       & \multicolumn{2}{c|}{UIT-HWDB-word} & \multicolumn{2}{c}{UIT-HWDB-line} \\ \hline
        Model          & CER               & WER                & CER               & WER                \\ \hline
        CRNN           & 9.93            & 20.76            & 31.23  & 100.00  \\
        BiCRNN         & 8.07            & 18.08            & 11.76           & 30.94            \\
        NRTR           & 8.25            & 21.31            & 49.72  & 91.70   \\
        TransformerOCR & \textbf{5.29}            & \textbf{10.37}            & \textbf{11.42}           & \textbf{21.67}            \\
        AED        & 6.70            & 14.68            & 12.53           & 35.31            \\ \hline
    \end{tabular}
\end{table}

We evaluated the baseline methods on UIT-HWDB-word and UIT-HWDB-line containing word-level and line-level handwriting images, respectively. Our results are shown in Table \ref{table:2}.

As shown in Table \ref{table:2}, both NRTR and CRNN obtained bad results on the UIT-HWDB-line dataset. This is because CRNN and NRTR were originally proposed to predict cropped-word scene text images, hence their backbone (CNN structure) were designed to be simple in order to enhance their performance \cite{shi2016end,8978180}. In contrast, Puigcerver et al. \cite{8269951} and Feng et al. \cite{feng2020scene} designed the BiCRNN and the TransformerOCR with every deep CNN structures. Therefore, these two models outperform NRTR and CRNN on recognizing both line-level and word-level handwriting images.

\subsection{Results analysis of the baseline methods}
As the results in our experiments, TransformerOCR achieved the best results among other methods on both the word-level and line-level dataset. Therefore, we focus on analyzing the common mistakes that TransformerOCR suffered.


\begin{figure*}[ht!]
    \centering
    \includegraphics[width=0.75\textwidth]{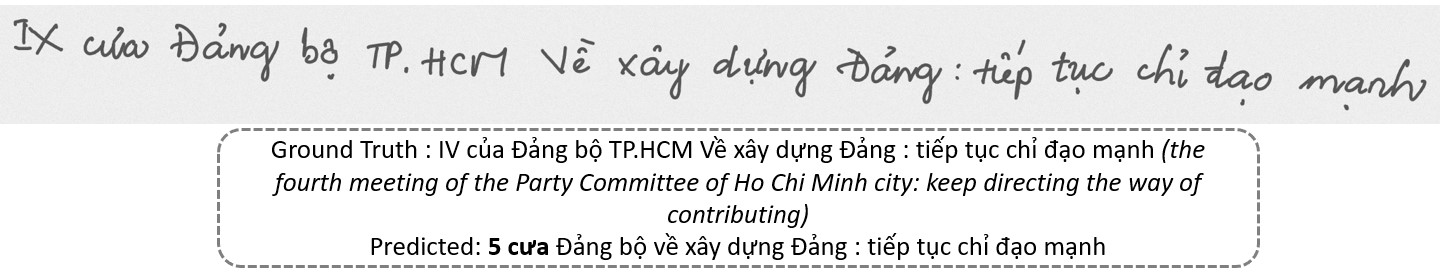}
    \caption{Wrong-number prediction because of the low frequency of numbers in training dataset.}
    \label{fig:error_transformer}
\end{figure*}

\begin{figure*}[ht!]
    \centering
    \includegraphics[width=0.75\textwidth]{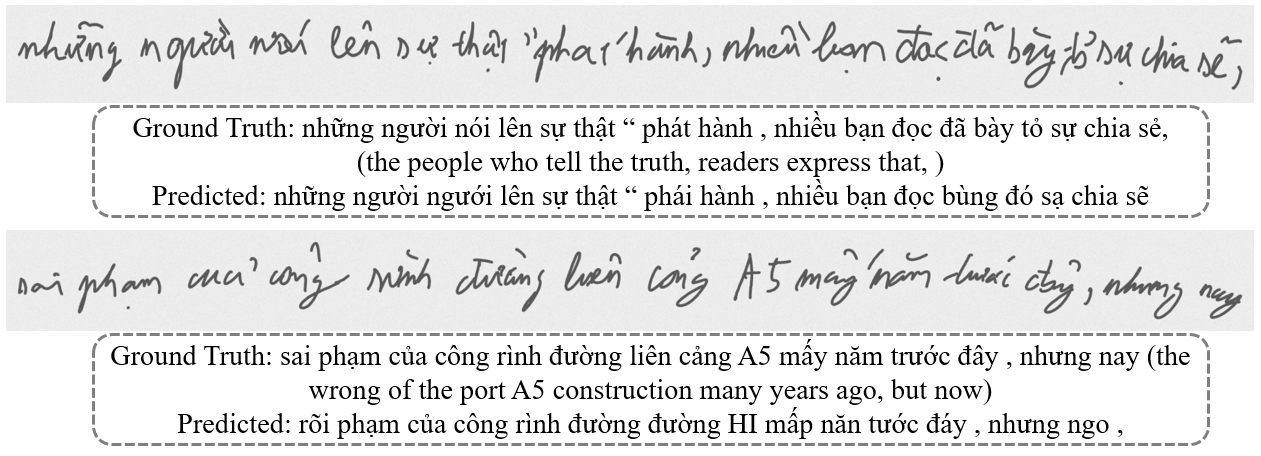}
    \caption{Examples for failed cases that the Transformer can not read scribbled characters.}
    \label{fig:wrong_transformer}
\end{figure*}

After observing the images in the test set and the respective prediction of the TransformerOCR, we conducted the two types of error of the TransformerOCR which are (1) the wrong prediction of number characters and (2) the misunderstanding of scrawled handwritten characters.

For the first mistake (Figure \ref{fig:error_transformer}), as the domain of the UIT-HWDB dataset is daily newspaper, the occurrence of numerical characters is lower than of alphabet characters,
which causes the model to not learn enough to make the correct decision when facing these numerical characters. Statistically, the frequency of numerical characters (including Roman numerical characters) in the UIT-HWDB-word train set is approximately 0.065 and in the UIT-HWDB-word test set is approximately 0.0066 (we calculated these frequencies on the word-level dataset because the line-level is the line-segmented version of paragraph-level, and the word-level is the word-segmented version of the line-level, which indicates the frequency of number characters in the word-level dataset is the frequency of them in the line-level and paragraph-level dataset). But as the low occurrence of number characters on the train set as well as the test set, this type of mistake is typically not the main factor to reduce the performance of the TransformerOCR method.

The last type of error is the misunderstanding of scribbled characters which is the main challenge of recognizing handwriting images as well as the main challenge of our dataset. As depicted in Figure \ref{fig:wrong_transformer}, TransformerOCR totally failed to achieve the acceptable prediction for images containing scrawled handwritten characters. For humans, we find these images are hard-to-read, but this does not mean reading these images is infeasible. The only way to read these words is to infer them 
together with meaning of the lines. Therefore we suggest the insight to improve the TransformerOCR or deep learning methods in general for this challenge is finding a way to exploit the meaning of the sentence to make models have better inference.
\section{Conclusion and Future Works}
We have introduced the Transferring method that synthetically forms a novel handwriting image dataset, which is useful for low-resource languages including Vietnamese. In addition, we presented a new benchmark for evaluating handwriting image recognition methods in Vietnamese constructed using this Transferring method. Finally, we have analyzed the confusion of state-of-the-art methods when performing on our dataset. We have identified that the appearance of number characters and badly scrawled text in the already linguistically complicated dataset are the causes behind the shortcomings in the predictions of the models. In the future, we continue to conduct experiments with various augmentation methods to tackle the shortage of number characters in our dataset. Moreover, we will analyze and research the way to effectively combine the handwriting recognition methods with pre-trained language models (like BERTology \cite{rogers2020primer}) to make them infer better when making predictions for scribbled handwriting images.

\bibliographystyle{IEEEtran}
\bibliography{anthology}

\end{document}